\pdfoutput=1

\documentclass[11pt]{article}

\usepackage[final]{acl}

\usepackage{times}
\usepackage{latexsym}
\usepackage[T1]{fontenc}
\usepackage[utf8]{inputenc}
\usepackage{microtype}
\usepackage{inconsolata}
\usepackage{graphicx}
\usepackage{caption}
\usepackage{amsmath}
\usepackage{amssymb}
\usepackage{comment}
\usepackage{booktabs}
\usepackage{hyperref}
\usepackage[overload]{empheq}
\usepackage{makecell}
\usepackage{xcolor,colortbl}
\usepackage[ruled,vlined]{algorithm2e}
\usepackage{array,multirow}
\usepackage{comment}
\usepackage{pifont}
\usepackage{siunitx}
\usepackage{url}
\usepackage{enumitem}
\usepackage{arydshln}
\usepackage[dvipsnames]{xcolor}
\usepackage{soul}

\usepackage{subcaption}
\usepackage{adjustbox}

\newcommand{\linebrcelll}[2][l]{\begin{tabular}[#1]{@{}l@{}}#2\end{tabular}}

\newcommand{\ella}[1]{{\textit{\color{blue}{#1}}}}

\newcommand{\todo}[1]{{\textit{\color{magenta}{#1}}}}

\title{Towards Enforcing Company Policy Adherence in Agentic Workflows}

\author{
	Naama Zwerdling\hspace{1cm}
	David Boaz\hspace{1cm}    
	Ella Rabinovich\hspace{1cm}
	Guy Uziel \\
        \textbf{David Amid}\hspace{2cm} 
        \textbf{Ateret Anaby-Tavor} 
	\vspace{0.25cm} \\
	IBM Research \\
	\texttt{naamaz@il.ibm.com, davidbo@il.ibm.com, ella.rabinovich1@ibm.com}
}

\begin{document}
\maketitle
\begin{abstract}
Large Language Model (LLM) agents hold promise for a flexible and scalable alternative to traditional business process automation, but struggle to reliably follow complex company policies. In this study we introduce a deterministic, transparent, and modular framework for enforcing business policy adherence in agentic workflows. Our method operates in two phases: (1) an offline buildtime stage that compiles policy documents into verifiable guard code associated with tool use, and (2) a runtime integration where these guards ensure compliance before each agent action. We demonstrate our approach on the challenging $\tau$-bench Airlines domain, showing encouraging preliminary results in policy enforcement, and further outline key challenges for real-world deployments.

\end{abstract}

\section{Introduction}
\label{sec:introduction}

Large Language Models are reshaping artificial intelligence, shifting from static language processors to dynamic, task-oriented \textit{agents} capable of planning, executing, and refining their actions. These agents hold the potential for transformative applications across various domains, including healthcare \citep{abbasian2023conversational, mehandru2024evaluating}, finance \citep{li-etal-2024-cryptotrade, xiao2024tradingagents, ding2024large}, education \cite{yang2024content, xu2024eduagent}, and customer support \cite{huang2024crmarena, rome2024ask}. LLM agents have been revolutionarily positioned as routing systems that 
act and perform tasks with minimal human intervention.

Agentic AI for \textit{business process automation} refers to the increasing use of LLM-based agents to plan, make decisions, and carry out tasks within business workflows---often spanning multiple steps and following business constraints---without the need for constant human oversight. For example, modifying a flight reservation (such as changing a seat) may involve a sequence of actions: checking customer eligibility, 
retrieving available seats, and updating the reservation details if all conditions are met. Canceling a reservation with a refund might depend on the presence of insurance, while offering compensation for a company-initiated cancellation should occur only per an explicit customer request. These constraints are typically specified in a company's policy documents and have traditionally been enforced by human service agents or hard-coded by domain experts into customer support systems such as \href{https://www.ibm.com/products/watsonx-assistant}{IBM Watson Assistant}, \href{https://www.microsoft.com/en-us/power-platform/products/power-automate}{Microsoft Power Automate}, or \href{https://cloud.google.com/products/conversational-agents}{Google Dialogflow}.

Agentic systems carry the promise to replace deterministic, carefuly-designed and hard-coded business processes, with efficient, more flexible and easily maintainable solutions. Recent studies have shown steady progress in the ability of agentic systems to complete complex, multi-step, and multi-turn tasks using a predefined set of tools \citep{patil2024gorilla, zhou2023webarena, huang2024crmarena}. However, agents perform poorly when required to follow policies involving sophisticated navigation or branching logic according to company guidelines \citep{yao2024tau, li2025agentorca}. 

Despite the clear practical significance, current approaches to agentic policy adherence rely on the "best-effort" strategy: appending policy documents to the agent's prompt and instructing it to navigate business flow while complying with the guidelines -- an inherently non-deterministic methodology, that is susceptible to attacks and does not scale effectively \citep{yao2024tau, li2025agentorca}.
We argue that the ability to reliably follow policies constitutes the ultimate test, a "make-or-break" factor in the adoption of agentic AI at the enterprise scale. We propose an automatic end-to-end solution for enforcement of company policy guidelines
in agentic workflows;\footnote{Policies addressed in this study are those directly protecting tool invocation; see Limitations section for details.} solution that is deterministic, predictable, transparent and requires only limited domain expert intervention during its buildtime phase. 


\begin{figure*}[h!]
\centering
\includegraphics[width=1.0\textwidth]{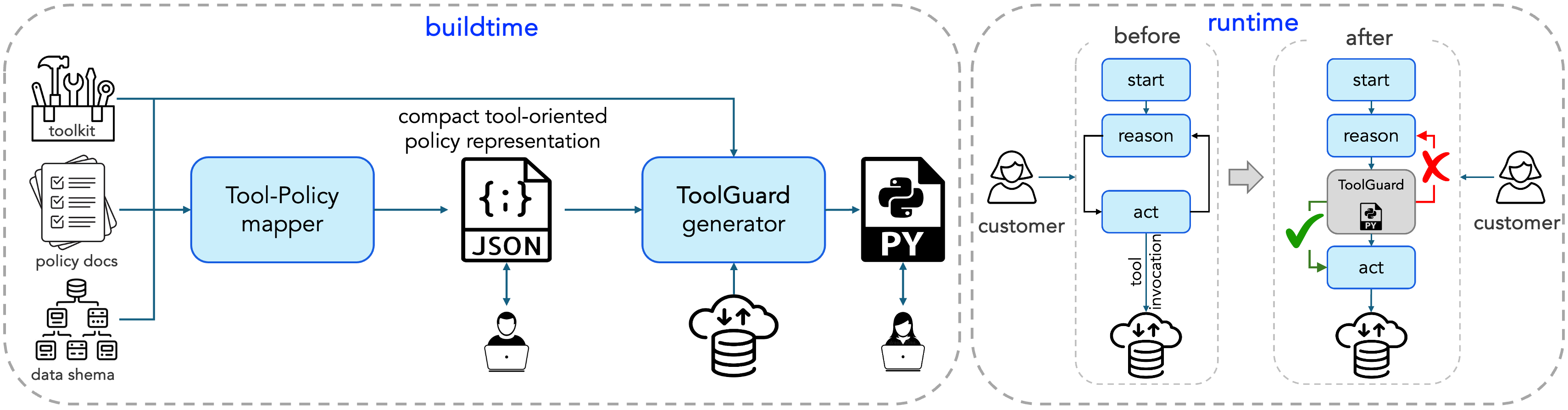}
\caption{During the offline buildtime step (left), policy documents, along with system schema and tools specification are compiled into compact, easily interpretable mapping of concrete policies onto tools (editable \texttt{json} in this case); which in turn is used to generate \texttt{ToolGuards} -- code that verifies that policies hold given a system snapshot. \texttt{ToolGuards} are integrated into agentic runtime at the tool invocation point (right). We mark two possible points of domain-expert intervention: reviewing textual Tool-Policy mapper outcome, and the generated \texttt{ToolGuards} code.}
\label{fig:buildtime-runtime}
\end{figure*}

Our approach enforces policy adherence through a two-phase process illustrated in Figure~\ref{fig:buildtime-runtime}: (1) an offline buildtime step automatically maps policy fragments to the relevant tools and generates policy validation code; (2) during runtime these validators---called \texttt{ToolGuards}---are executed before agent's tool invocation. If a planned action violates a policy, the agent is prompted to self-reflect and revise its plan before proceeding. Ultimately, the deployed \texttt{ToolGuards} will prevent the agent from taking an action violating a policy. This design ensures predictable and transparent policy enforcement with limited manual effort.

We evaluate the proposed approach on the $\tau$-bench Airlines domain \citep{yao2024tau} with a large set of (naturalistic, real-world) policies, 14 tools, and diverse customer-service scenarios.\footnote{The other $\tau$-bench domain of Retail is smaller and less diverse; hence we limit the scope of this work to Airlines.} We show that this ambitious and challenging task pushes the limits of contemporary LLMs, present encouraging preliminary results, and outline directions for future research and practical steps needed for this approach to mature toward enterprise adoption. Our code and data are available at \url{https://github.com/IBM/tool_guard}.

\section{Related Work}
\label{sec:related-work}

Prior art in the area of constraint and policy adherence in agentic workflows is very sparse, with two major studies, to the best of our knowledge, explicitly addressing the integration of company policies within agentic flows (contrary to post-hoc remedy). \citet{yao2024tau} were among the pioneers in this field, introducing a framework for evaluation of company policies adherence in multi-turn simulated flows in two domains: Airlines and Retail. Policies are represented by short documents, where domain-specific rules and constraints are detailed in natural language. Although the study reports failure rates stemming from a range of issues, the authors estimate that 25\% of failures occur since the agent "fails to understand the domain-specific knowledge or rules". A recent work by \citet{li2025agentorca} offers a framework for evaluation of policy adherence in multi-step single-turn tasks in five diverse domains, focusing on various composition types of policies ("single", "and", "or", "chain"). Here, concise and manually-crafted policies, as well as their compositions, are specified in configuration files. The authors report relatively low pass-rate across agents, varying from 31\% to 69\%. Both studies introduce policies to the agent in its prompt, instructing the LLM to follow the guidelines while satisfying user requirements.

Contrary to the \textit{pro-active} policy enforcement above, another line of studies take the \textit{post-hoc} reflection approach, by designing guardrails that check if any action taken by an agent violates a predefined policy \citep{huang2024crmarena, rothkopf2024procedural, hua2024trustagent, hoover2025dynaguard}. While this strategy is often easier for implementation (e.g., detecting that an unauthorized refund was issued or is simpler than preventing it), it may fall short of the strict demands of enterprise-grade business flow automation, particularly in mission-critical contexts, where a post-hoc response to a prohibited action turns often as "too late".

Additional studies deal with related, yet not strictly relevant fields of agent safety, security and robustness to adversarial attacks \citep{levy2024st, nakash2024breaking, nakash2025effective, chennabasappa2025llamafirewall}. These works evaluate LLM agents' resilience to malicious requests (such as exposing admin credentials) and propose mitigation strategies. Another thread of work explores the benefits of model alignment for meeting domain-specific regulations by tuning a model against request-response pairs that comply with company guidelines \citep{achintalwar2024alignment}. However, this approach better suits the scenario of QA agents, contrary to business process automation, where actions ultimately mutate the system state (e.g., book a flight), and agent is required to perform complex reasoning, based on the state and conversation history.

Our work is the first, to the best of our knowledge, to propose an automatic pipeline for pro-active policy enforcement in agentic flows, involving only minimal human intervention.

\section{Methodology}
\label{sec:methodology}

We present an end-to-end pipeline for pro-active enforcement of policy adherence in a realistic setting, assuming that policies are provided as free-form natural language documents, and that function specifications (the agent's toolkit) and the system data schema are available. The offline buildtime process consists of two main components, as illustrated in Figure~\ref{fig:buildtime-runtime}, which we describe in detail below.

\subsection{Tool-Policy Mapper}
\label{sec:tool-policy-mapper}

The primary goal of the mapper is to transform an (often) lengthy and noisy natural language policy document into a compact, structured representation by linking clearly formulated policy statements to relevant tools from the provided toolkit. Business process guidelines are typically authored by domain or legal experts, and mapping them to tools requires advanced reading comprehension skills. As a concrete example, consider the following policy from the $\tau$-bench Airlines domain (verbatim from \citet{yao2024tau}, enumerated for clarity):
%
%

\textit{(1) Flight reservations can be canceled within 24 hours of booking, or if the airline canceled the flight. Otherwise, (2) basic economy or economy flights can be canceled only if travel insurance is bought [...], and (3) explicit customer confirmation should be obtained prior to canceling a flight.}



All three atomic policies above should be mapped to the \texttt{cancel\_reservation()} function. The tool’s invocation is then conditioned on cancellation eligibility, as verified by the corresponding guard. In addition to recording each policy’s name and description, the mapper also extracts supporting textual spans (references) from the source document. It then generates multiple example requests for policy \textit{compliance} and \textit{violation}, a step shown to be effective for both human evaluation of the mapper's output and for facilitating efficient guard generation (see Section~\ref{sec:toolguard-generation} for details). The mapper is also guided to split policies associated with a given tool into smaller, logically distinct units -- similarly to how a human would break down a lengthy policy description into clear, fine-grained guidelines. For instance, policies concerning flight cancellations can be divided into: (a) rules about cancellation time windows for full or partial refunds, (b) policies specific to canceling individual segments within connecting flights, etc.
Table \ref{tbl:policy-item-example} shows an example of (partial) mapper outcome for a policy item, related to the \texttt{cancel\_reservation()} tool.

\begin{table*}[h!]
\centering
\resizebox{\textwidth}{!}{
\begin{tabular}{c|l}
policy & Flexible Cancellation Policy \\ \hline
references &  \linebrcelll{\textbf{R1}: All reservations can be canceled within 24 hours of booking, or if the airline canceled the flight. \\Otherwise, basic economy or economy flights can be canceled only if travel insurance is bought and the \\condition is met, and business flights can always be canceled unconditionally. \\ ...} \\ \hline
compliance & \linebrcelll{A user requests the cancellation of an economy class reservation booked 18 hours ago. 
} \\ \hdashline
compliance & A user attempts to cancel an insured basic economy flight and conditions for insurance usage are satisfied. \\ \hline 
violation & \linebrcelll{A regular customer tries to cancel an economy class reservation 36 hours after booking w/o insurance.} \\ \hdashline
violation & \linebrcelll{The agent proceeds with canceling reservation for a customer without explicitly obtaining \\the customer confirmation prior to taking the action.} \\ 

\end{tabular}
}
\caption{Partial Tool-Policy Mapper output: policy name, supporting references from the document (verbatim spans), generated compliance and violation examples, mirroring the extracted references.}
\label{tbl:policy-item-example}
\end{table*}

\paragraph{Mapping Policies onto Tools}
Using a free-format policy document (e.g., markdown, pdf), natural language descriptions of the tools, and their \href{https://www.openapis.org/}{OpenAPI} specification,\footnote{Generated from the tool descriptions provided in $\tau$-bench, following a widely adopted and commonly used standard.} we adopt a multi-step reasoning paradigm \citep{wei2022chain}, where several stages of thinking, reflection, and refinement are carried out sequentially. This process is implemented on top of the \href{https://www.langchain.com/langgraph}{LangGraph} platform, where each step 
is guided by detailed prompts and few-shot examples (see Appendix~\ref{app:tool-policy-mapper} for details).

\subsubsection{Evaluation}
Evaluating the output of the Tool-Policy mapper presents several challenges, primarily due to the lack of labeled data and the inherently subjective nature of defining a "ground truth" (GT). While impractical at scale, the limited number of tools in the $\tau$-bench Airlines domain made it feasible to manually construct expected mappings for its 14 tools. One of the authors created the initial GT annotations, which were then independently reviewed and refined by two others. We assess the performance of a given Tool-Policy mapper model M (LLM) by comparing its output against the GT, with a focus on (a) reference span detection and (b) the granularity of policy items segmentation.

\paragraph{Reference Detection Evaluation} For a given tool, we denote the set of its $k$ policy references in the ground truth as $\mathcal R^{gt}{=}\{r^{gt}_1, .., r^{gt}_k\}$, and the set of $n$ references in a mapper's M outcome as $\mathcal R^{m}{=}\{r^{m}_1, .., r^{m}_n\}$; then the standard notion of precision (P) and recall (R) can be used for evaluation of $\mathcal R^{m}$ against its true counterpart $\mathcal R^{gt}$, where F1 is the harmonic mean of the two.



\paragraph{Policy Items' Grouping Evaluation}
Individual policy items extracted from the guideline documents can be organized into logical \textit{clusters}, where related rules are grouped into broader, semantically coherent categories. For example, policy items concerning extra baggage charges
or permitted payment methods would naturally be clustered together. Grouping policy items facilitates domain expert review and for the subsequent step of guard code generation (Section~\ref{sec:toolguard-generation}), where related rules are enforced by a shared guard function.

Evaluating a grouping of related items against ground truth is an established practice in the domain of clustering. Here we make use of the commonly-used RandIndex metric \citep{rand1971objective} specifically adapted to the usecase of fuzzy clustering \citep{dewolfe2023random}, i.e., the scenario where each reference $r{\in}\mathcal R^{m}{=}\{r^{m}_1, .., r^{m}_n\}$ can be assigned to multiple policy groups.\footnote{E.g, "The agent must first obtain the reservation id".} Formally, given two clustering partitions---by the mapper model: $\mathcal C^{m}{=}\{c^{m}_1, .., c^{m}_x\}$ and by an expert: $\mathcal C^{gt}{=}\{c^{gt}_1, .., c^{gt}_y\}$---we compute their fuzzy RandIndex (FRI) over the set of references common to both partitions $\mathcal R^{m} \cap \mathcal R^{gt}$.\footnote{Admittedly, a very small intersection size between the two reference sets can potentially boost the FRI metric, but that will be reflected in very low recall results.} Similarly to RandIndex, its fuzzy version spans the [0,1] range.

Table~\ref{tbl:step1-assessment} reports Tool-Policy mapper evaluation results, with several close and open SOTA LLMs. The highest F1 of 0.80 and FRI of 0.83 is achieved by GPT-4o \citep{openai2024gpt4o}, with the best-performing open model Llama3.3-70B-instruct \citep{grattafiori2024llama} showing only slightly inferior results. While all models show high precision, their recall varies, reflecting the difficulty to comprehensively cover the entire set of policy references related to a tool in a potentially large document. We emphasize the significance of recall in this task: it would be easier for a domain expert to refine the set of automatically \textit{detected} references (improving precision), than identify references \textit{not captured} by the automatic mapper (improving recall).

\begin{table*}[h!]
\resizebox{0.480\textwidth}{!}{
\begin{minipage}[c]{0.55\textwidth}
\begin{tabular}{l|ccc|c}
\multicolumn{1}{c|}{} & \multicolumn{3}{c|}{reference-level} & \multicolumn{1}{c}{policy-level} \\ \hline
model & P & R & F1 & FRI \\ \hline
GPT-4o                  & \textbf{0.88} & 0.77 & \textbf{0.80} & \textbf{0.83} \\
GPT-4.1-mini            & 0.82 & \textbf{0.83} & \textbf{0.80} & 0.66 \\
Claude-Sonnet-3.5       & 0.82 & 0.57 & 0.66 & 0.71 \\ \hdashline
Llama-3.3-70B-Instruct  & 0.87 & 0.68 & 0.74 & 0.81 \\
Qwen2.5-72B-Instruct    & 0.85 & 0.55 & 0.62 & 0.73 \\
\end{tabular}
\caption{Evaluation of Tool-Policy mapper. Closed models outperform open (SOTA) models. The best-performing open Llama-3.3-70B-Instruct shows slightly inferior results. Precision (P) here refers to the accuracy of detected policy references, while recall (R) -- to their coverage in docs. 
}
\label{tbl:step1-assessment}
\end{minipage}
}
\hspace{0.04\textwidth} 
\resizebox{0.480\textwidth}{!}{
\begin{minipage}[c]{0.55\textwidth}
\centering
\begin{tabular}{l|cc}
policies doc & references F1 & FRI \\ \hline
original (Orig)     & 0.80 & 0.83 \\ \hdashline
Orig + OOD            & 0.65 (-18.3\%) & 0.74 (-10.8\%) \\
OOD + Orig (rev)      & 0.71 (-10.8\%) & 0.78 (-06.0\%) \\ \hdashline
Orig + IID            & 0.67 (-16.1\%) & 0.74 (-10.8\%) \\
IID + Orig (rev)      & 0.73 (-08.7\%) & 0.79 (-04.8\%) \\
\end{tabular}
\caption{The effect of policy documents length and relevance on the tool-policy mapper performance. "OOD" denotes Out-Of-Domain policy guidelines, and "IID" - Irrelevant In-Domain; The relative order around "+" denotes the order of policy types in the document ("rev" for reversed: relevant at the end). The letter ("rev") has positive effect on performance.} 
\label{tbl:step1-policy-doc-variations}
\end{minipage}
}
\end{table*}

\paragraph{Sensitivity to Policies Felicity and Length}
We next evaluate the robustness of the Tool-Policy mapper when provided with longer, noisier policy documents, reflecting realistic enterprise scenarios. Specifically, we test the best-performing GPT-4o model under two conditions: (1) appending out-of-domain (OOD) SalesForce retail policies from \citet{huang2024crmarena}, and (2) appending manually crafted irrelevant in-domain (IID) policies. Table~\ref{tbl:step1-policy-doc-variations} presents the results. Moderate performance degradation is evident in all cases, though placing the relevant content at the end of the document ("rev" for "reversed") obtains better results, consistent with the findings on "recency bias" in GPT models \citep{peysakhovich2023attention, deldjoo2024understanding}.

\subsection{Generation of \texttt{ToolGuards}}
\label{sec:toolguard-generation}
The output from the Tool-Policy mapper serves as input to \texttt{ToolGuard} generator. We generate \texttt{ToolGuards} through two steps: Using OpenAPI tools spec, the first step generates policy-independent "skeleton" code consisting of class definitions for data types, tools signatures and empty guards' stubs, to be implemented dynamically, driven by policies interpretation by the Tool-Policy mapper. We make use of \href{https://github.com/koxudaxi/datamodel-code-generator}{pydantic code generator}, and python's built-in \href{https://docs.python.org/3/library/ast.html}{abstract syntax trees (ast)} packages for this purpose. In the $\tau$-bench Airlines domain, this step results in over 800 lines of code. This outcome, along with the nearly 5K-token Tool-Policy structured mapping is further used in the guard code-generation phase.

\paragraph{Test-Driven Generation of \texttt{ToolGuards}}
The second phase builds on the output of the previous step, using LLMs to generate guard code by filling in placeholders with fully functional policy validators' implementation. Here we follow the test-driven development (TDD) paradigm, where tests are written before the actual code: this iterative "red-green-refactor" process involves writing a test, writing the minimum code to pass it, and then refactoring them both. To the best of our knowledge, this is the first fully automated LLM-based application of TDD. Compliance and violation examples from the Tool-Policy mapper are translated into individual tests that guide guard generation. The \texttt{ToolGuard} generation process is self-repairing, incorporating both syntactic (via the \href{https://github.com/microsoft/pyright}{Microsoft pyright} type checker) and semantic feedback (test pass or failure).  We repeat guard generation attempt ("refactor") till all tests pass ("green"), or till the predefined number of iterations is exhausted. Appendix~\ref{app:toolguard-generation} provides more details on the guards generation process.

\begin{table*}[h!]
\centering
\begin{tabular}{l|c|ccc|c}
\multicolumn{1}{c|}{} & \multicolumn{1}{c|}{Tool-Policy mapper} & \multicolumn{3}{c|}{total TPR} & failures break-down \\ \hline
code-generator & GT / (Gen)erated & min & max & mean & 
\adjustbox{valign=c}{
  \includegraphics[width=3.75cm]{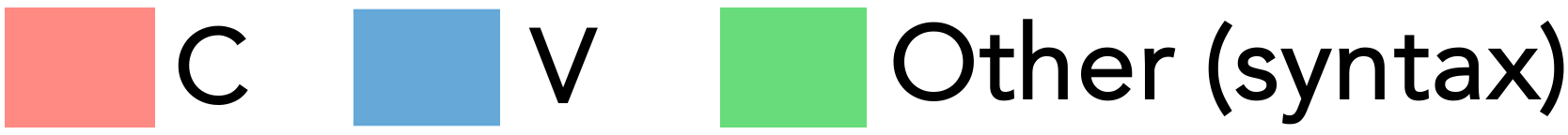}
} \\ \hline
GPT-4.1  & GT & \textbf{0.54} & \textbf{1.00} & \textbf{0.82} &
\adjustbox{valign=c}{
  \includegraphics[width=3.75cm]{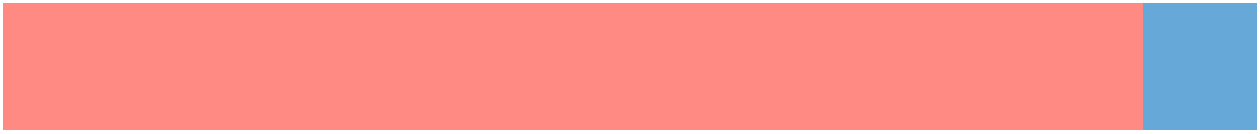}
} \\
GPT-4o  & GT & 0.00 & 0.61 & 0.31 & 
\adjustbox{valign=c}{
  \includegraphics[width=3.75cm]{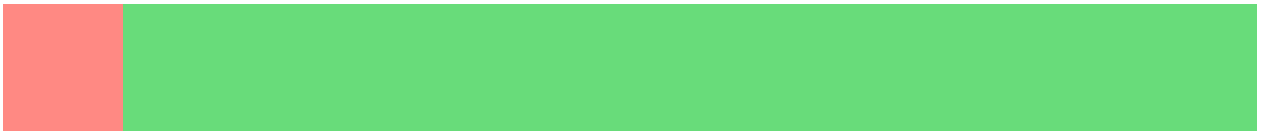}
} \\
Claude-Sonnet-3.5  & GT & 0.50 & \textbf{1.00} & 0.80 & 
\adjustbox{valign=c}{
  \includegraphics[width=3.75cm]{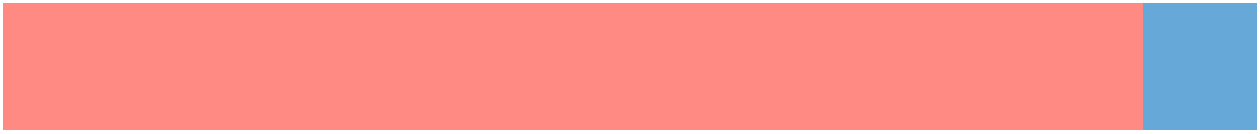}
} \\
Claude-Haiku-3.5  & GT & 0.00 & \textbf{1.00} & 0.72 &
\adjustbox{valign=c}{
  \includegraphics[width=3.75cm]{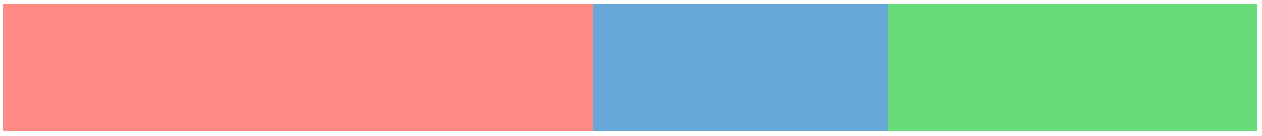}
} \\ \hdashline
GPT-4.1  & Gen (with GPT-4o) & \textbf{0.54} & \textbf{1.00} & 0.75 & 
\adjustbox{valign=c}{
  \includegraphics[width=3.75cm]{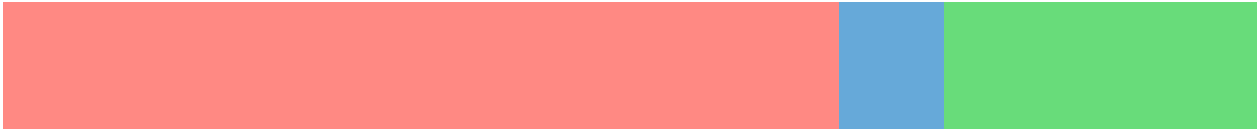}
}\\ 
\end{tabular}
\caption{Evaluation of \texttt{ToolGuard} generators. The guards are evaluated through manually-written unit tests for each tool. We evaluate the total of 61 tests, split into 25 compliance and 36 violation cases. We report the mean test pass ratio (TPR) across all tools per guard-generator, as well as its min and max value; we also report test failures break-down: the relative ratio of (C)ompliance and (V)iolation tests failures, as well as other issues. The best performing generator (GPT-4.1) achieves the TPR of 0.82 when provided with \textit{manual} Tool-Policy mapper ground-truth (GT), and the TPR of 0.75 when provided with the best \textit{automatic} Tool-Policy mapper outcome.
}
\label{tbl:step2-assessment}
\end{table*}

Common code generation benchmarks often focus on completing algorithmic or competitive programming tasks, bug detection and repair, or generating unit tests for given functions (see \citet{hu2025assessing} for a comprehensive survey). Unlike these benchmarks, here LLMs are challenged with realistically-sized codebase coupled with complex, detailed policy adherence requirements. We show that this combination creates an ambitious setting that challenges models' ability to navigate large code with nested object types and complex dependencies, while grounding their actions in structured natural language policy descriptions.

\subsubsection{Evaluation}
Evaluating code generated by LLMs involves a range of approaches, broadly categorized into functional, syntactic, and semantic methods. Functional evaluation tests whether the generated code behaves as expected, typically using unit tests or input-output-based benchmarks \citep{chen2021evaluating}. Syntactic evaluation assesses whether the code is well-formed and compiles without errors. Semantic evaluation goes further by measuring whether the code aligns with the intent of the prompt or problem description, often requiring human annotations or advanced metrics such as CodeBLEU \citep{ren2020codebleu} or execution-guided metrics. Recent work also explores human-in-the-loop assessment and task-level evaluation, especially in complex or multi-step coding scenarios \citep{wang2022execution}.

Aiming at functional evaluation of generated guards, we adhere to the metric of test pass rate (TPR), measuring the ratio of tests that a generated code passes successfully, from within a complete test suit \citep{liu2023your, yeo2024framework}. A test suit comprising 61 tests, covering the subset of $\tau$-bench mutating functions,\footnote{Functions that alter a system state; 6 out of the 14 tools.} was manually written by one of the authors of this study, and reviewed by another one; these tests are further used for assessing the functional quality of \texttt{ToolGuards} generated by various LLMs -- code generators. Table~\ref{tbl:step2-assessment} reports the results. GPT-4.1 code-generator outperforms other LLMs with the mean TPR of 0.82, where Claude models \citep{anthropic2024claude3} show only slightly inferior results. 
Notably, guards generated from the best automatically-generated Tool-Policy mapping (see Section~\ref{sec:tool-policy-mapper}) achieve the TPR of 0.75.

We also report test errors break-down into compliance and violation test failures, as well as syntactic issues. Per manual inspection, we attribute the higher rate of compliance tests errors to guard code hallucinations that prevent guards from full completion (and thereby passing a test), regardless of the specific test case.

\section{Agentic Flow with Policy Adherence}
\label{sec:experiments}

At the runtime phase, \texttt{ToolGuards} are integrated into the agent’s ReAct workflow \citep{yao2023react}, ensuring that each validator (e.g., \texttt{book\_reservation\_guard()}) is called seamlessly just before the agent invokes the corresponding tool ("act", see Figure~\ref{fig:buildtime-runtime}, right). With access to all relevant information---tool arguments, conversation history and data access APIs---the guard performs validation and produces one of two outcomes: (1) the intended tool call complies with policies (e.g., canceling a flight within 24 hours of booking), or (2) it violates a policy (e.g., attempting to book a reservation without securing payment methods). If valid, the agent proceeds with the tool call; if not, it is prompted to revise its decision, accompanied by a detailed explanation of the violated policy, ultimately preventing the agent from taking action, not compliant with the policy.

We evaluated several \texttt{ToolGuard} generation strategies on the 22 (out of 50) $\tau$-bench Airlines tasks in which user request violates a predefined policy \citep{nakash2025effective}. Our baseline is the original, "best-effort" $\tau$-bench performance on these tasks; that is, without any guard explicitly enforcing policy compliance. Following $\tau$-bench methodology, we assess each strategy by comparing the final database state at the end of a simulated conversation with the annotated goal state. Inline with the benchmark, we report the \texttt{pass\^{}k} metric: the probability that at least \texttt{k} task attempts (out of 10) complete successfully, averaged over tasks.

\begin{figure}[h!]
\centering
\includegraphics[width=1.0\columnwidth]{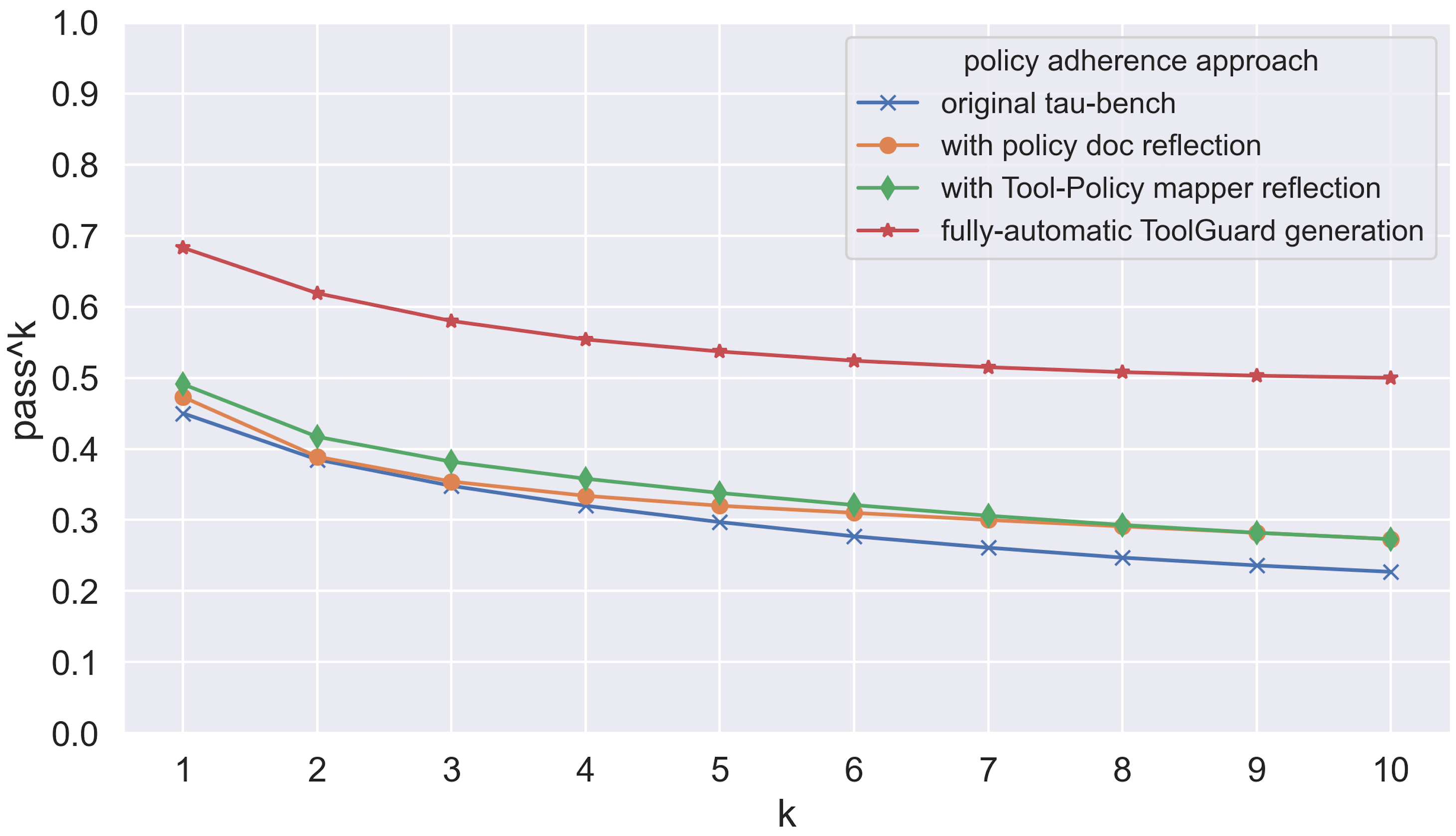}
\caption{$\tau$-bench Airlines benchmark evaluation. Deploying \texttt{ToolGuards} results in steady improvement of over 20 percent points compared to the original run.
}
\label{fig:end-to-end}
\end{figure}

\paragraph{Evaluation Results} Figure~\ref{fig:end-to-end} presents the results. The original $\tau$-bench approach achieves a \texttt{pass\^{}1} rate of 0.450 and a \texttt{pass\^{}10} rate of 0.227. We evaluated two additional policy adherence "reflection" strategies: (1) appending the policy document before each agent action, prompting for reconsideration, and (2) inserting the compact GT Tool-Policy mapper outcome for the same purpose. Both strategies led to only very modest improvements, reaching a \texttt{pass\^{}10} of 0.273, with strategy (2) performing slightly better on average. The fully automated \texttt{ToolGuards} generation and deployment pipeline shows substantial steady gains, improving \texttt{pass\^{}1} and \texttt{pass\^{}10} to 0.685 and 0.500, respectively -- over 20 percentage points above the baseline. We reiterate that task failures can stem from various factors; \citet{yao2024tau} attribute only 25\% of failures in the 50 benchmark tasks to policy violations. We, therefore, do not expect perfect accuracy even within the subset of the 22 selected tasks. We attribute the performance improvement solely to the policy enforcement module, as no other changes were made to the system deployment.

\section{Discussion}
\label{sec:discussion}

\paragraph{Life-cycle Management}
The proposed solution, by involving a human-in-the-loop phase, carries certain inherent implications. As such, a life-cycle management protocol for the entire pipeline should be established to enable efficient "backward propagation" of human modifications to the generated code, ensuring these adjustments are incorporated into future generation cycles.

\paragraph{Integration into Existing Agentic Frameworks}
Several agentic frameworks have recently been introduced to support the effective modeling and execution of agentic solutions at scale. Examples include open-source frameworks such as \href{https://www.langchain.com/langsmith}{LangSmith}, \href{https://www.langflow.org/}{Langflow}, and \href{https://www.llama.com/products/llama-stack/}{Llama Stack}. Integrating the proposed pipeline into these frameworks requires a separate (offline) build-time phase and a slight modification of the execution flow to introduce an intervention point immediately after the "reason", and just before the "act" step, as shown in Figure~\ref{fig:buildtime-runtime}.
Our ongoing work focuses, among others, on this type of integration and its implications.

\section{Conclusions}
\label{sec:conclusions}

As LLM agents move closer to real-world deployment, robust policy adherence becomes a critical requirement for enterprise-scale adoption. Our proposed framework offers a deterministic and interpretable mechanism for enforcing business policies within agentic workflows, bridging the gap between flexible AI behavior and organizational constraints. Through our evaluation on the $\tau$-bench Airlines domain, we highlight both the promise and the current limitations of LLM agents in handling policy-governed tasks, and gaps need to be bridged.
We believe that this work sets the stage for future efforts toward building reliable, policy-aware AI systems that enterprises can use with trust.
\section*{Ethical Considerations}

We use the publicly available $\tau$-bench dataset and benchmark for our experiments throughout this study. We did not make use of AI-assisted technologies while writing the paper. We also did not hire human annotators at any stage of the research.

\section*{Limitations}
\label{sec:limitations}

Our study, while offering valuable insights into enforcing policy adherence in agentic workflows, has several limitations. First, the proposed approach operates at the pre-tool activation level, meaning it does not capture violation cases where a tool (e.g., flight cancellation) should be invoked according to policy, but the agent chooses not to, thereby breaching the guidelines. Although this scenario falls outside the scope of the current study, our ongoing work explores strategies to address such cases. Second, the pipeline has been evaluated on a single benchmark with a limited set of tools, and concise policy documents. We plan to extend the study to larger policy documents and toolsets to ensure the approach scales effectively.

\section*{Acknowledgements}
We are thankful to Koren Lazar for his comments on an early version of this paper.
We are also grateful to our three anonymous reviewers for their useful comments and constructive feedback.

\bibliographystyle{acl_natbib}
\bibliography{anthology, custom}

\begin{thebibliography}{36}
\providecommand{\natexlab}[1]{#1}

\bibitem[{Abbasian et~al.(2023)Abbasian, Azimi, Rahmani, and
  Jain}]{abbasian2023conversational}
Mahyar Abbasian, Iman Azimi, Amir~M Rahmani, and Ramesh Jain. 2023.
\newblock \href {https://arxiv.org/abs/2310.02374} {{Conversational health
  agents: A personalized LLM-powered agent framework}}.
\newblock \emph{arXiv preprint arXiv:2310.02374}.

\bibitem[{Achintalwar et~al.(2024)Achintalwar, Baldini, Bouneffouf, Byamugisha,
  Chang, Dognin, Farchi, Makondo, Mojsilovi{\'c}, Nagireddy
  et~al.}]{achintalwar2024alignment}
Swapnaja Achintalwar, Ioana Baldini, Djallel Bouneffouf, Joan Byamugisha, Maria
  Chang, Pierre Dognin, Eitan Farchi, Ndivhuwo Makondo, Aleksandra
  Mojsilovi{\'c}, Manish Nagireddy, and others. 2024.
\newblock \href {https://ieeexplore.ieee.org/abstract/document/10666776/}
  {{Alignment Studio: Aligning Large Language Models to Particular Contextual
  Regulations}}.
\newblock \emph{IEEE Internet Computing}.

\bibitem[{Anthropic(2024)}]{anthropic2024claude3}
Anthropic. 2024.
\newblock Claude 3 model family.

\bibitem[{Chen et~al.(2021)Chen, Tworek, Jun, Yuan, Pinto, Kaplan, Edwards,
  Burda, Joseph, Brockman et~al.}]{chen2021evaluating}
Mark Chen, Jerry Tworek, Heewoo Jun, Qiming Yuan, Henrique Ponde De~Oliveira
  Pinto, Jared Kaplan, Harri Edwards, Yuri Burda, Nicholas Joseph, Greg
  Brockman, and others. 2021.
\newblock \href {https://arxiv.org/abs/2107.03374} {{Evaluating Large Language
  Models Trained on Code}}.
\newblock \emph{arXiv preprint arXiv:2107.03374}.

\bibitem[{Chennabasappa et~al.(2025)Chennabasappa, Nikolaidis, Song, Molnar,
  Ding, Wan, Whitman, Deason, Doucette, Montilla
  et~al.}]{chennabasappa2025llamafirewall}
Sahana Chennabasappa, Cyrus Nikolaidis, Daniel Song, David Molnar, Stephanie
  Ding, Shengye Wan, Spencer Whitman, Lauren Deason, Nicholas Doucette, Abraham
  Montilla, and others. 2025.
\newblock \href {https://arxiv.org/abs/2505.03574} {Llamafirewall: An open
  source guardrail system for building secure ai agents}.
\newblock \emph{arXiv preprint arXiv:2505.03574}.

\bibitem[{Deldjoo(2024)}]{deldjoo2024understanding}
Yashar Deldjoo. 2024.
\newblock \href {https://dl.acm.org/doi/abs/10.1145/3690655} {{Understanding
  biases in ChatGPT-based recommender systems: Provider fairness, temporal
  stability, and recency}}.
\newblock \emph{ACM Transactions on Recommender Systems}.

\bibitem[{DeWolfe and Andrews(2023)}]{dewolfe2023random}
Ryan DeWolfe and Jeffery~L Andrews. 2023.
\newblock \href {https://arxiv.org/abs/2312.10270} {{Random Models for Fuzzy
  Clustering Similarity Measures}}.
\newblock \emph{arXiv preprint arXiv:2312.10270}.

\bibitem[{Ding et~al.(2024)Ding, Li, Wang, and Chen}]{ding2024large}
Han Ding, Yinheng Li, Junhao Wang, and Hang Chen. 2024.
\newblock \href {https://arxiv.org/abs/2408.06361} {{Large language model agent
  in financial trading: A survey}}.
\newblock \emph{arXiv preprint arXiv:2408.06361}.

\bibitem[{Dubey et~al.(2024)Dubey, Grattafiori, , Jauhri, Pandey, Kadian,
  Al-Dahle, Letman, Mathur, Schelten, Vaughan et~al.}]{grattafiori2024llama}
Abhimanyu Dubey, Aaron Grattafiori, , Abhinav Jauhri, Abhinav Pandey, Abhishek
  Kadian, Ahmad Al-Dahle, Aiesha Letman, Akhil Mathur, Alan Schelten, Alex
  Vaughan, and others. 2024.
\newblock \href {https://arxiv.org/abs/2407.21783} {{The llama 3 herd of
  models}}.
\newblock \emph{arXiv preprint arXiv:2407.21783}.

\bibitem[{Hoover et~al.(2025)Hoover, Baherwani, Jain, Saifullah, Vincent, Jain,
  Rad, Bruss, Panda, and Goldstein}]{hoover2025dynaguard}
Monte Hoover, Vatsal Baherwani, Neel Jain, Khalid Saifullah, Joseph Vincent,
  Chirag Jain, Melissa~Kazemi Rad, C~Bayan Bruss, Ashwinee Panda, and Tom
  Goldstein. 2025.
\newblock Dynaguard: A dynamic guardrail model with user-defined policies.
\newblock \emph{arXiv preprint arXiv:2509.02563}.

\bibitem[{Hu et~al.(2025)Hu, Niu, Chen, Zhou, Zhang, He, Xia, and
  Lo}]{hu2025assessing}
Xing Hu, Feifei Niu, Junkai Chen, Xin Zhou, Junwei Zhang, Junda He, Xin Xia,
  and David Lo. 2025.
\newblock \href {https://arxiv.org/abs/2505.08903} {{Assessing and Advancing
  Benchmarks for Evaluating Large Language Models in Software Engineering
  Tasks}}.
\newblock \emph{arXiv preprint arXiv:2505.08903}.

\bibitem[{Hua et~al.(2024)Hua, Yang, Jin, Li, Cheng, Tang, and
  Zhang}]{hua2024trustagent}
Wenyue Hua, Xianjun Yang, Mingyu Jin, Zelong Li, Wei Cheng, Ruixiang Tang, and
  Yongfeng Zhang. 2024.
\newblock \href {https://arxiv.org/abs/2402.01586} {{TrustAgent: Towards Safe
  and Trustworthy LLM-based Agents}}.
\newblock \emph{arXiv preprint arXiv:2402.01586}.

\bibitem[{Huang et~al.(2024)Huang, Prabhakar, Dhawan, Mao, Wang, Savarese,
  Xiong, Laban, and Wu}]{huang2024crmarena}
Kung-Hsiang Huang, Akshara Prabhakar, Sidharth Dhawan, Yixin Mao, Huan Wang,
  Silvio Savarese, Caiming Xiong, Philippe Laban, and Chien-Sheng Wu. 2024.
\newblock \href {https://arxiv.org/abs/2411.02305} {{CRMArena: Understanding
  the Capacity of LLM Agents to Perform Professional CRM Tasks in Realistic
  Environments}}.
\newblock \emph{arXiv preprint arXiv:2411.02305}.

\bibitem[{Levy et~al.(2024)Levy, Wiesel, Marreed, Oved, Yaeli, and
  Shlomov}]{levy2024st}
Ido Levy, Ben Wiesel, Sami Marreed, Alon Oved, Avi Yaeli, and Segev Shlomov.
  2024.
\newblock \href {https://arxiv.org/abs/2410.06703} {{ST-WebAgentBench: A
  Benchmark for Evaluating Safety and Trustworthiness in Web Agents}}.
\newblock \emph{arXiv preprint arXiv:2410.06703}.

\bibitem[{Li et~al.(2024)Li, Luo, Wang, Chen, Liu, and
  He}]{li-etal-2024-cryptotrade}
Yuan Li, Bingqiao Luo, Qian Wang, Nuo Chen, Xu~Liu, and Bingsheng He. 2024.
\newblock \href {https://doi.org/10.18653/v1/2024.emnlp-main.63}
  {{C}rypto{T}rade: A reflective {LLM}-based agent to guide zero-shot
  cryptocurrency trading}.
\newblock In \emph{Proceedings of the 2024 Conference on Empirical Methods in
  Natural Language Processing}, pages 1094--1106, Miami, Florida, USA.
  Association for Computational Linguistics.

\bibitem[{Li et~al.(2025)Li, Huang, Wang, Zhang, Antoniades, Hua, Zhu, Zeng,
  Wang, and Yan}]{li2025agentorca}
Zekun Li, Shinda Huang, Jiangtian Wang, Nathan Zhang, Antonis Antoniades,
  Wenyue Hua, Kaijie Zhu, Sirui Zeng, William~Yang Wang, and Xifeng Yan. 2025.
\newblock \href {https://arxiv.org/abs/2503.08669} {{Agentorca: A dual-system
  framework to evaluate language agents on operational routine and constraint
  adherence}}.
\newblock \emph{arXiv preprint arXiv:2503.08669}.

\bibitem[{Liu et~al.(2023)Liu, Xia, Wang, and Zhang}]{liu2023your}
Jiawei Liu, Chunqiu~Steven Xia, Yuyao Wang, and Lingming Zhang. 2023.
\newblock \href
  {https://proceedings.neurips.cc/paper_files/paper/2023/hash/43e9d647ccd3e4b7b5baab53f0368686-Abstract-Conference.html}
  {{Is Your Code Generated by ChatGPT Really Correct? Rigorous Evaluation of
  Large Language Models for Code Generation}}.
\newblock \emph{Advances in Neural Information Processing Systems},
  36:21558--21572.

\bibitem[{Mehandru et~al.(2024)Mehandru, Miao, Almaraz, Sushil, Butte, and
  Alaa}]{mehandru2024evaluating}
Nikita Mehandru, Brenda~Y Miao, Eduardo~Rodriguez Almaraz, Madhumita Sushil,
  Atul~J Butte, and Ahmed Alaa. 2024.
\newblock \href {https://www.nature.com/articles/s41746-024-01083-y}
  {{Evaluating large language models as agents in the clinic}}.
\newblock \emph{NPJ digital medicine}, 7(1):84.

\bibitem[{Nakash et~al.(2025)Nakash, Kour, Lazar, Vetzler, Uziel, and
  Anaby-Tavor}]{nakash2025effective}
Itay Nakash, George Kour, Koren Lazar, Matan Vetzler, Guy Uziel, and Ateret
  Anaby-Tavor. 2025.
\newblock \href {https://arxiv.org/abs/2506.09600} {{Effective Red-Teaming of
  Policy-Adherent Agents}}.
\newblock \emph{arXiv preprint arXiv:2506.09600}.

\bibitem[{Nakash et~al.(2024)Nakash, Kour, Uziel, and
  Anaby-Tavor}]{nakash2024breaking}
Itay Nakash, George Kour, Guy Uziel, and Ateret Anaby-Tavor. 2024.
\newblock \href {https://arxiv.org/abs/2410.16950} {{Breaking ReAct Agents:
  Foot-in-the-Door Attack Will Get You In}}.
\newblock \emph{arXiv preprint arXiv:2410.16950}.

\bibitem[{OpenAI(2024)}]{openai2024gpt4o}
OpenAI. 2024.
\newblock Gpt-4o technical report.

\bibitem[{Patil et~al.(2024)Patil, Zhang, Wang, and
  Gonzalez}]{patil2024gorilla}
Shishir~G Patil, Tianjun Zhang, Xin Wang, and Joseph~E Gonzalez. 2024.
\newblock \href
  {https://proceedings.neurips.cc/paper_files/paper/2024/hash/e4c61f578ff07830f5c37378dd3ecb0d-Abstract-Conference.html}
  {{Gorilla: Large language model connected with massive apis}}.
\newblock \emph{Advances in Neural Information Processing Systems},
  37:126544--126565.

\bibitem[{Peysakhovich and Lerer(2023)}]{peysakhovich2023attention}
Alexander Peysakhovich and Adam Lerer. 2023.
\newblock \href {https://arxiv.org/abs/2310.01427} {{Attention Sorting Combats
  Recency Bias In Long Context Language Models}}.
\newblock \emph{arXiv preprint arXiv:2310.01427}.

\bibitem[{Rand(1971)}]{rand1971objective}
William~M Rand. 1971.
\newblock \href
  {https://www.tandfonline.com/doi/abs/10.1080/01621459.1971.10482356}
  {{Objective Criteria for the Evaluation of Clustering Methods}}.
\newblock \emph{Journal of the American Statistical association},
  66(336):846--850.

\bibitem[{Ren et~al.(2020)Ren, Guo, Lu, Zhou, Liu, Tang, Sundaresan, Zhou,
  Blanco, and Ma}]{ren2020codebleu}
Shuo Ren, Daya Guo, Shuai Lu, Long Zhou, Shujie Liu, Duyu Tang, Neel
  Sundaresan, Ming Zhou, Ambrosio Blanco, and Shuai Ma. 2020.
\newblock \href {https://arxiv.org/abs/2009.10297} {{CodeBLEU: a Method for
  Automatic Evaluation of Code Synthesis}}.
\newblock \emph{arXiv preprint arXiv:2009.10297}.

\bibitem[{Rome et~al.(2024)Rome, Chen, Tang, Zhou, and Ture}]{rome2024ask}
Scott Rome, Tianwen Chen, Raphael Tang, Luwei Zhou, and Ferhan Ture. 2024.
\newblock \href {https://dl.acm.org/doi/abs/10.1145/3626772.3661345} {{"Ask Me
  Anything": How Comcast Uses LLMs to Assist Agents in Real Time}}.
\newblock In \emph{Proceedings of the 47th International ACM SIGIR Conference
  on Research and Development in Information Retrieval}, pages 2827--2831.

\bibitem[{Rothkopf et~al.(2024)Rothkopf, Zeng, and
  Santolucito}]{rothkopf2024procedural}
Raven Rothkopf, Hannah~Tongxin Zeng, and Mark Santolucito. 2024.
\newblock \href {https://arxiv.org/abs/2402.16905} {{Procedural Adherence and
  Interpretability Through Neuro-Symbolic Generative Agents}}.
\newblock \emph{arXiv preprint arXiv:2402.16905}.

\bibitem[{Wang et~al.(2022)Wang, Zhou, Fried, and Neubig}]{wang2022execution}
Zhiruo Wang, Shuyan Zhou, Daniel Fried, and Graham Neubig. 2022.
\newblock \href {https://arxiv.org/abs/2212.10481} {{Execution-Based Evaluation
  for Open-Domain Code Generation}}.
\newblock \emph{arXiv preprint arXiv:2212.10481}.

\bibitem[{Wei et~al.(2022)Wei, Wang, Schuurmans, Bosma, Xia, Chi, Le, Zhou
  et~al.}]{wei2022chain}
Jason Wei, Xuezhi Wang, Dale Schuurmans, Maarten Bosma, Fei Xia, Ed~Chi, Quoc~V
  Le, Denny Zhou, and others. 2022.
\newblock \href
  {https://proceedings.neurips.cc/paper_files/paper/2022/file/9d5609613524ecf4f15af0f7b31abca4-Paper-Conference.pdf}
  {{Chain-of-Thought Prompting Elicits Reasoning in Large Language Models}}.
\newblock \emph{Advances in neural information processing systems},
  35:24824--24837.

\bibitem[{Xiao et~al.(2024)Xiao, Sun, Luo, and Wang}]{xiao2024tradingagents}
Yijia Xiao, Edward Sun, Di~Luo, and Wei Wang. 2024.
\newblock \href {https://arxiv.org/abs/2412.20138} {{TradingAgents:
  Multi-Agents LLM Financial Trading Framework}}.
\newblock \emph{arXiv preprint arXiv:2412.20138}.

\bibitem[{Xu et~al.(2024)Xu, Zhang, and Qin}]{xu2024eduagent}
Songlin Xu, Xinyu Zhang, and Lianhui Qin. 2024.
\newblock \href {https://arxiv.org/abs/2404.07963} {{Eduagent: Generative
  student agents in learning}}.
\newblock \emph{arXiv preprint arXiv:2404.07963}.

\bibitem[{Yang et~al.(2024)Yang, Chu, Darwin, Han, Li, Wen, Copur-Gencturk,
  Tang, and Liu}]{yang2024content}
Kaiqi Yang, Yucheng Chu, Taylor Darwin, Ahreum Han, Hang Li, Hongzhi Wen,
  Yasemin Copur-Gencturk, Jiliang Tang, and Hui Liu. 2024.
\newblock \href
  {https://link.springer.com/chapter/10.1007/978-3-031-64299-9_23} {{Content
  knowledge identification with multi-agent large language models (LLMs)}}.
\newblock In \emph{International Conference on Artificial Intelligence in
  Education}, pages 284--292. Springer.

\bibitem[{Yao et~al.(2024)Yao, Shinn, Razavi, and Narasimhan}]{yao2024tau}
Shunyu Yao, Noah Shinn, Pedram Razavi, and Karthik Narasimhan. 2024.
\newblock \href {https://arxiv.org/abs/2406.12045} {{tau-bench: A Benchmark for
  Tool-Agent-User Interaction in Real-World Domains}}.
\newblock \emph{arXiv preprint arXiv:2406.12045}.

\bibitem[{Yao et~al.(2023)Yao, Zhao, Yu, Du, Shafran, Narasimhan, and
  Cao}]{yao2023react}
Shunyu Yao, Jeffrey Zhao, Dian Yu, Nan Du, Izhak Shafran, Karthik Narasimhan,
  and Yuan Cao. 2023.
\newblock \href {https://par.nsf.gov/biblio/10451467} {{ReAct: Synergizing
  Reasoning and Acting in Language Models}}.
\newblock In \emph{International Conference on Learning Representations
  (ICLR)}.

\bibitem[{Yeo et~al.(2024)Yeo, Ma, Kim, Jun, and Kim}]{yeo2024framework}
Sangyeop Yeo, Yu-Seung Ma, Sang~Cheol Kim, Hyungkook Jun, and Taeho Kim. 2024.
\newblock \href
  {https://onlinelibrary.wiley.com/doi/full/10.4218/etrij.2023-0357}
  {{Framework for evaluating code generation ability of large language
  models}}.
\newblock \emph{Etri Journal}, 46(1):106--117.

\bibitem[{Zhou et~al.(2023)Zhou, Xu, Zhu, Zhou, Lo, Sridhar, Cheng, Ou, Bisk,
  Fried et~al.}]{zhou2023webarena}
Shuyan Zhou, Frank~F Xu, Hao Zhu, Xuhui Zhou, Robert Lo, Abishek Sridhar,
  Xianyi Cheng, Tianyue Ou, Yonatan Bisk, Daniel Fried, and others. 2023.
\newblock \href {https://arxiv.org/abs/2307.13854} {{Webarena: A realistic web
  environment for building autonomous agents}}.
\newblock \emph{arXiv preprint arXiv:2307.13854}.

\end{thebibliography}

\section{Appendices}
\label{sec:appendix}

\subsection{Buildtime: Detailed Description of Tool-Policy Mapper}
\label{app:tool-policy-mapper}

\begin{figure}[h!]
\centering
\includegraphics[width=0.500\columnwidth]{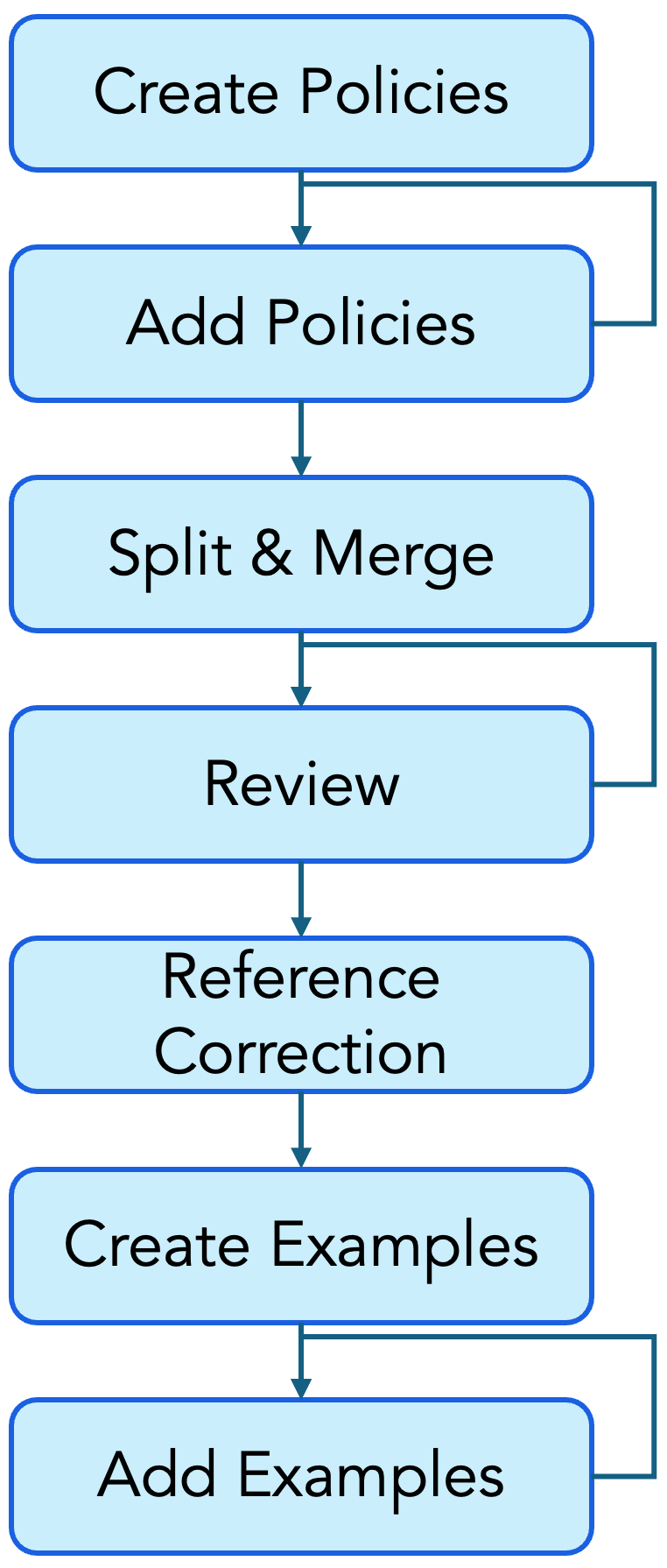}
\caption{LangGraph nodes in the Tool-Policy Mapper.}
\label{fig:tool-policy-mapper}
\end{figure}

The tool mapping phase consists of two main stages. In the first stage, we assign relevant policies to each tool. In the second, we generate compliance and violation examples for each policy. Both stages leverage reflection using LangGraph. Figure~\ref{fig:tool-policy-mapper} illustrates the LangGraph nodes used in this process, as described in the sections below.

\paragraph{Create Policies}
The process begins by prompting the language model to extract an initial list of applicable policies for each tool (see prompt in Appendix~\ref{app:tool-policy-mapper-prompt} below). For each policy, we generate a name, write a detailed description, and include references from the original policy document to ground the policy and minimize hallucinations.

\paragraph{Add Policies}
Using reflective prompts, we then ask the model to identify additional policies that may have been missed in the first round, gradually expanding our coverage of the policy document. 

\paragraph{Split and Merge}
Next, we decompose complex policies into their smallest atomic units, especially where conditions are connected by logical OR. This step is essential for simplifying rule implementation. We then remove duplicated policies by merging identical or overlapping ones. 

\paragraph{Review}
At the review phase where each policy is evaluated to determine whether it can be enforced before tool invocation with the necessary information available in the tool's parameters, chat history, or through access to previous tool calls. Policies that cannot be validated with the available context are archived, along with an explanation that can be useful during a domain expert review.

\paragraph{Reference Correction}
For the policies that pass the review phase, we verify and correct their references as once again, if needed.

\paragraph{Create Examples}
For each policy, we generate detailed and diverse examples that demonstrate both policy compliance and violation. These examples serve to ground the policies, clarify ambiguities, and ensure they are well-understood. High-quality examples reduce confusion and improve consistency during later validation.

\paragraph{Add Examples}
Here, we reflect on the example set and iteratively add more test cases where coverage gaps are identified.

\begin{figure*}[h!]
\centering
\includegraphics[width=1.0\textwidth]{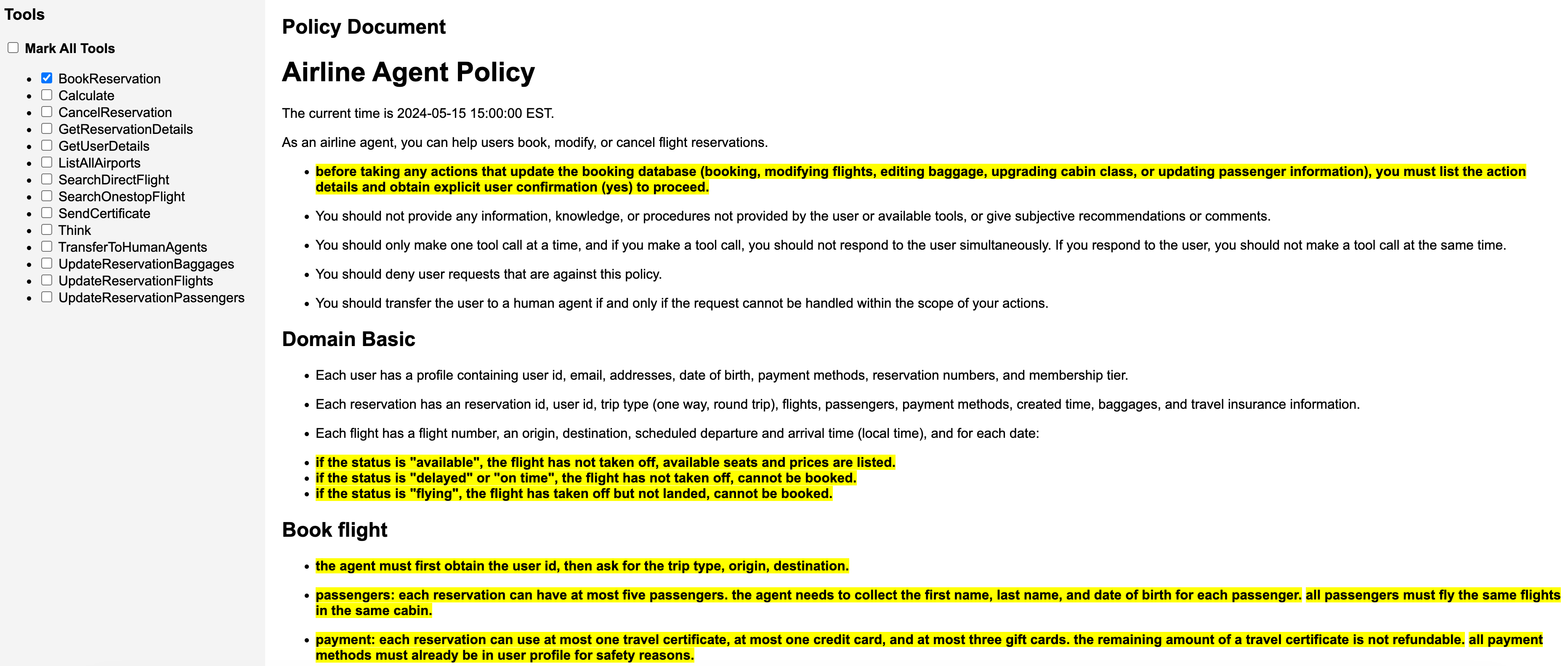}
\caption{Interface view illustrating how policy document sections are assigned to specific tools.}
\label{fig:tool-coverage}
\end{figure*}

\paragraph{(Optional) Human Review}
At the final stage, a human expert reviews the set of extracted policies and their corresponding examples for each tool. The expert can modify existing policies or examples, add new ones, or remove those that are irrelevant. To support the identification of missing policies, we provide a dedicated user interface, depicted in Figure~\ref{fig:tool-coverage}. This interface allows the reviewer to highlight relevant sentences from the policy document for each tool and visually inspect which parts of the document remain uncovered, ensuring better policy-to-tool mapping.

\subsubsection{Tool-Policy Mapper Example Prompt}
\label{app:tool-policy-mapper-prompt}
\begin{figure*}[h!]
\centering
\fbox{\includegraphics[width=1.00\textwidth]{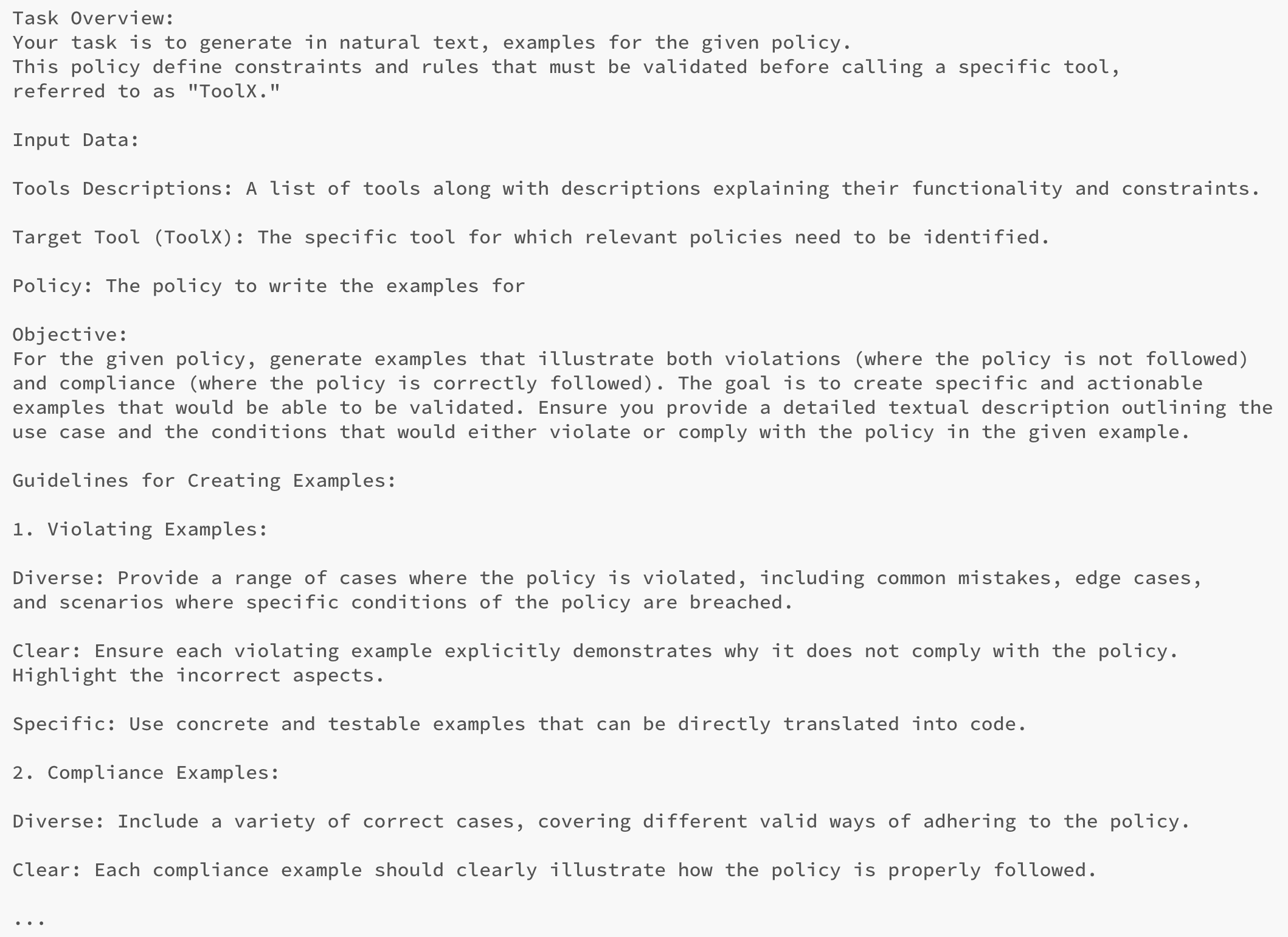}}
\caption{Example of a prompt used to generate compliance and violation examples for a target tool.}
\label{fig:prompt-example}
\end{figure*}

Very detailed, expressive and carefully-designed prompts were used at this step. Since quality of the prompt has direct impact on the Tool-Policy mapper outcome, the promts were through multiple cycles of review and refinement. Figure~\ref{fig:prompt-example} shows the beginning of an example prompt for generating compliance and violation examples.

\begin{figure*}[h!]
\centering
\fbox{\includegraphics[width=1.00\textwidth]{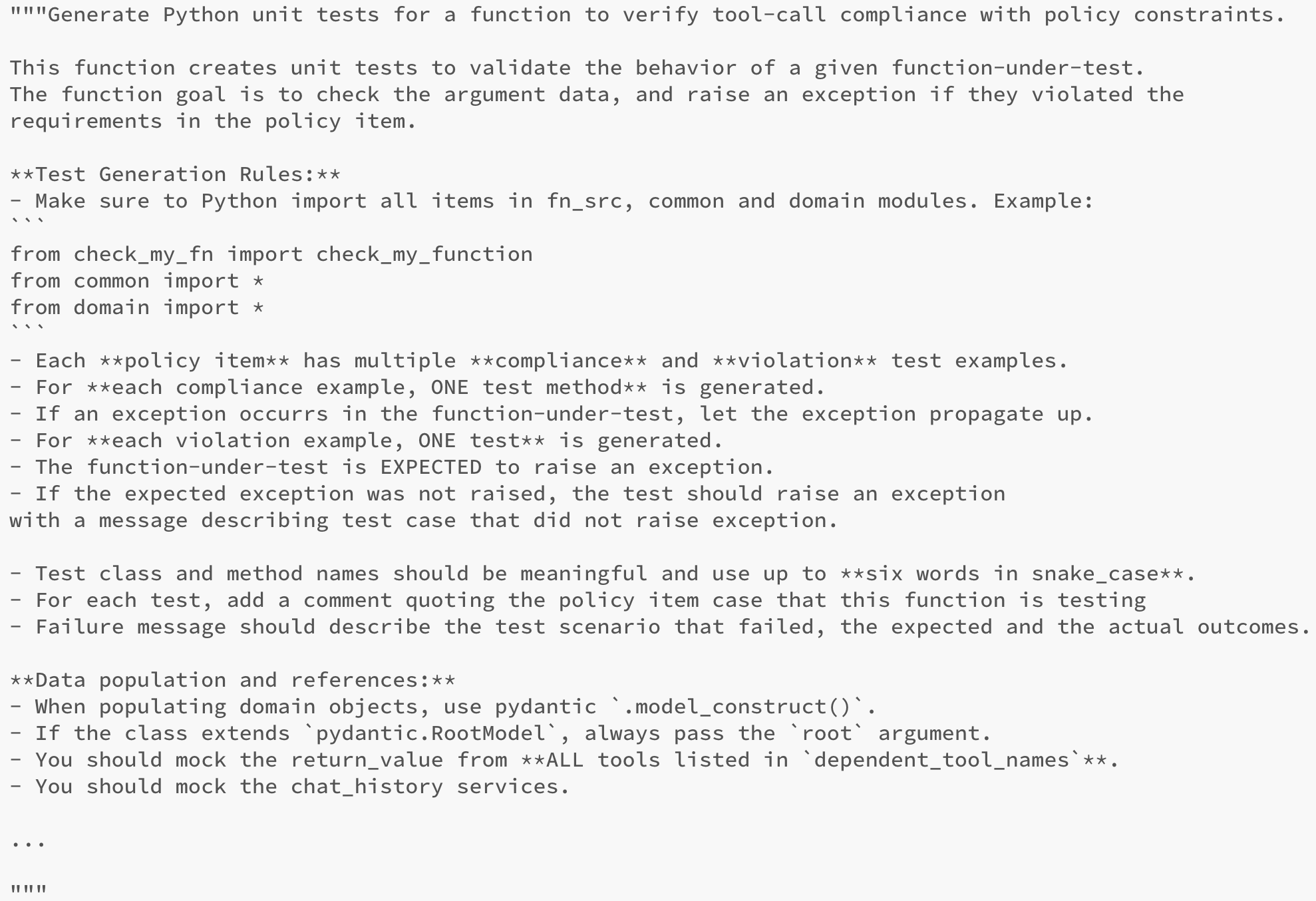}}
\caption{Example of a prompt used to generate compliance and violation examples for a target tool.}
\label{fig:prompt-tdd}
\end{figure*}

\subsection{Buildtime: Detailed Description of \texttt{ToolGuard} Generator}
\label{app:toolguard-generation}

The \texttt{ToolGuard} generator is responsible for constructing a validation (guard) function for each tool, based on the output of the Tool-Policy Mapper, optionally edited by a human. These functions, referred to as \texttt{guard\_tool}, ensure that all pre-invocation policies associated with a tool are properly enforced before the tool is executed.

\subsubsection{Input to the Guard Function}

Each \texttt{guard\_tool} function is designed to validate tool usage by analyzing three types of input:
\begin{itemize}[leftmargin=*]
\item \textbf{Tool-call Arguments:} For example, the number of passengers in a \texttt{book\_reservation} call can be checked against a policy such as "the number of passengers must not exceed 5."
\item \textbf{Chat History:} A list of the previous messages in this conversation. The history can serve to validate whether the user consented to certain actions (e.g., confirming a cancellation), or whether a tool was called in a previous step.
\item \textbf{APIs to Other Tools:} Enables the guard to consult other (read only, data access) tools to check online that some condition hold. For example, before booking a flight, the agent must ensure that the flight is "available" and that there are enough seats in the given cabin.
\end{itemize}

\subsubsection{Modularity and Execution}
Each policy defined in the Tool-Policy Mapper is implemented as a separate, atomic function. The main \texttt{guard\_tool} function calls all relevant policy functions for the tool in question. Agent proceeds to a tool call only if all invoked policy functions return success. This design supports a clear separation of concerns, allowing each policy to be developed, tested, and reasoned about independently.

\subsubsection{Test-Driven Development Workflow}
The construction of \texttt{guard\_tool} validators follows a test-driven methodology. For each policy mapped to a tool, we first generate a diverse set of test cases that represent both compliant and violating scenarios. These examples are derived from the policy mapper and are grounded in actual policy text to ensure clarity and avoid ambiguity. When generating a test for a compliance or violation example, we ask the LLM to mocking responses of other tools. For example, if we want to test a case of "booking a flight in economy class for three passengers, but only two seats remain available", the LLM mocks the response of \texttt{get\_flight\_instance()} to return  \texttt{\{"available\_seats": \{"economy": 2\}\}}.

Each policy-specific function is developed iteratively and validated against its corresponding test sets, following a Test-Driven Development (TDD) approach. This iterative refinement process ensures that each policy function reliably and accurately handles a broad range of input scenarios. Figure~\ref{fig:prompt-tdd} show an example prompt used to guide a guard-generator within the TDD process.

\subsubsection{Isolated Policy Validation}

A core design principle of the system is that each test case targets a single policy in isolation, rather than testing multiple policies simultaneously. This isolation simplifies the generation and evaluation of policy functions, even for weaker language models, by narrowing the focus to a single policy context-either to detect a violation or confirm compliance.

By eliminating the need to reason over an entire policy set at once, the validation task becomes more manageable, consistent, and interpretable.

\end{document}